\documentclass{article}
\usepackage{ijcai16}
\usepackage{graphicx}
\usepackage{amssymb}
\usepackage{amsmath}
\usepackage{bm}
\usepackage[linesnumbered,ruled, vlined]{algorithm2e}
\usepackage{multirow}
\usepackage{float}
\usepackage{times}
\newcommand{\specialcell}[2][c]{%
  \begin{tabular}[#1]{@{}c@{}}#2\end{tabular}}

\title{Resolving Spatial-Time Conflicts In A Set Of Any-angle Or Angle-constrained Grid Paths\thanks{This work was partially supported by RFBR, research project No. 15-37-20893.}}
\author{Konstantin Yakovlev\\ 
Federal Research Center\\ ``Computer Science and Control''\\
of Russian Academy of Sciences, \\Moscow, Russia \\
yakovlev@isa.ru \And Anton Andreychuk\\
Peoples' Friendship University of Russia,\\ Moscow, Russia \\
Federal Research Center \\``Computer Science and Control''\\
of Russian Academy of Sciences, \\Moscow, Russia \\
andreychuk@isa.ru }

\begin{document}

\hyphenpenalty=4000 
\maketitle
\begin{abstract}
We study the multi-agent path finding problem (MAPF) for a group of agents which are allowed to move into arbitrary directions on a 2D square grid. We focus on centralized conflict resolution for independently computed  plans. We propose an algorithm that eliminates conflicts by using local re-planning and introducing time offsets to the execution of paths by different agents. Experimental results show that the algorithm can find high quality conflict-free solutions at low computational cost.
\end{abstract}

\section{Introduction}

Square grids are simple yet informative models of both artificial (typically appearing in video games) and physical (involved in robotics) 2D environments used for path planning \cite{yap2002}. Typically in grid path finding an agent is presumed to move from one traversable (unoccupied) cell to one of it's 8 adjacent neighbors. Sometimes diagonal moves are prohibited restricting agent's moves to only the 4 cardinal directions. Various path planning methods have been proposed so far to find non-conflicting paths for multiple agents who move on a grid in such way: HCA* \cite{silver2005}, OD+ID \cite{standley2011}, MAPP \cite{wangbotea2011}, CBS \cite{sharon2015} etc. Some of these methods were intially developed to be applied to grid worlds (like MAPP), while others (like CBS) can apply to arbitrary graphs (grids including).
	
At the same time the limitations of 8 (or 4) connected grids have led to increased popularity of any angle path-finding. In any-angle path finding an agent is allowed to move into arbitrary direction and a valid move is represented by a line segment, {\it e.g.} a section, whose endpoints are tied to the distinct grid elements (either centers or corners of the cells) and which does not intersect any untraversable cell. Any-angle path planning algorithms like Theta* \cite{nash2007}, Anya \cite{harabor2013} etc. tend to find shorter and more realistic looking paths for numerous practical applications. Moreover for many applications, for example -- mobile robotics, it can be beneficial to search for angle-constrained paths, {\it e.g.} grid paths comprised of the sections having the property the angle between two consecutive sections does not exceed the predefined threshold \cite{kim2014}, \cite{yakovlev2015}.

To the best of authors knowledge multi-agent any-angle (or angle-constrained) path finding problem has received little attention from the community (although more general, continuous problems have been studied recently \cite{bento2015}). This work aims at filling this gap. We present preliminary results aimed at creating robust angle-constrained MAPF algorithm suitable for state-of-the-art intelligent control systems for mobile robots \cite{emelyanov2016}. Proposed algorithm falls into the category of sub-optimal decentralized planners (like MAPP, WHCA* etc.) meaning that agents paths are first generated independently then checked for conflicts, after which the conflicts are solved in an iterative fashion to produce the final solution -- a set of the conflict-free paths. No theoretical proofs of correctness are available yet, but the results of experimental evaluation involving realistic navigation scenarios suggests the effectiveness of the proposed algorithm.

\section{Problem Setting}

\subsection{Motivation}

We are motivated by the following scenario. Consider n agents that are unmanned multirotor aerial vehicles performing nap-of-the earth flight in urban environment at constant speed and at fixed height (speed and height are the same for all UAVs). All the vehicles start at their intial positions on the ground and are provided with the goal locations. When the mission starts the UAVs take off reaching the predefined height (some can stay on the ground and take off later) and perform the flights without colliding with the obstacles and each other.  The agents must maintain constant forwards velocity while in flight and can stop and land only at their target locations (thus preventing other agents from colliding with them).  The goal of the MAPF planner is to produce a conflict-free set of the individual solutions. The cost of the overall solution is the mission execution time.

The spatial path for a single agent is represented by a sequence of traversable line segments. We are additionally interested in such individual paths that are implicitly compatible with the dynamic constraints of the UAVs {\it e.g.} angle-constrained paths (ac-paths). An ac-path is a path having the property that an angle between any two consecu-tive segments does not exceed fixed predefined threshold. Ac-paths are smooth and do not contain sharp turns and thus can be easily followed by the UAVs performing flight at a constant speed.

\subsection{Formal Statement}

Consider a set on {\it n} agents placed on a finite grid of cells A that can be represented as a matrix  
$\text{\it A}_{\text{\it H}\times \text{\it W}}$$=$$\{ \text{\it a}_\text{\it ij} \}$, where {\it i},{\it j} -- are cell position indexes (coordinates) and {\it H}, {\it W} -- are grid dimensions. Using cell indexes a distance between any two grid cells can be calculated using Euclidean metrics.

We adopt center-based grid notation and assume that agents locations are tied to the centers of the cells. Each cell is either traversable or un-traversable for the agents and {\it n} start and goal locations, $\{\text {\it s}^{(1)},  \ldots , \text{\it s}^{(\text{\it n})}\}, \{\text{\it g}^{(1)},  \ldots , \text{\it g}^{(\text{\it n})}\}$, are given, {\it e.g.} corresponding traversable grid cells are given.

A line connecting centers of two grid cells (startpoint and endpoint) is called a section and denoted as $\text{\it e}$$=$$\langle\text{\it a}_\text{\it ij}, \text{\it a}_\text{\it kl}\rangle $$=$$\langle\text{\it sp}(\text{\it e}), \text{\it ep}(\text{\it e})\rangle$.  The section is considered to be traversable if the line-of-sight function invoked on section's endpoints returns true. Following the most common approach we use Bresehnam algorithm \cite{bresenham1965} to construct line-of-sight. The length of the section is a distance between it's startpoint and endpoint: $\text{\it len}(\text{\it e})$$=$$\text{\it dist}(\text{\it a}_\text{\it ij}, \text{\it a}_\text{\it kl})$. A section whose length is close to some fixed integer $\Delta$, {\it e.g. round$($len$($e$))$}$=$$\Delta$, is called $\Delta$-section.

Given two adjacent sections $\text{\it e}_\text{\it 1}$$=$$\langle \text{\it a}_{\text{\it ij}}, \text{\it a}_{\text{\it kl}}\rangle, \text{\it e}_2$$=$$\langle \text{\it a}_{\text{\it kl}}, \text{\it a}_{\text{\it vw}}\rangle$ an angle of alteration is the angle between the vectors which coordinates are $(${\it k $-$ i}, {\it l $-$ j}$)$ and $(${\it v $-$ k}, {\it w $-$ l}$)$ respectively. This angle is denoted as $\alpha(\text{\it e}_1, \text{\it e}_2)$ and it's value is denoted as $|\alpha(\text{\it e}_1, \text{\it e}_2)|$.

The path between two distinct traversable cells {\it s} (start cell) and {\it g} (goal cell) is a sequence of traversable adjacent sections such that the first section starts with s and the last ends with $\text{\it g}: \pi=\pi(\text{\it s, g})=\{\text{\it e}_1,\ldots, \text{\it e}_\text{\it v}\}, \text{\it e}_1=\langle \text{\it s}, \text{\it a}_{\text{\it ij}}\rangle, \text{\it e}_\text{\it v}=\langle \text{\it a}_{\text{\it kl}}, \text{\it g}\rangle,\newline \text{\it a}_{\text{\it ij}}, \text{\it a}_{\text{\it kl}} \in \text{\it A}_{\text{\it H}\times \text{\it W}}$. The length of the path is the sum lengths of its sections: $\text{\it len}(\pi)$$=$$\text{\it len}(\text{\it e}_1)$$+$$\ldots$$+$$\text{\it len}(\text{\it e}_\text{\it v})$.

Given a path $\pi$$=$$\{\text{\it e}_1,\ldots, \text{\it e}_\text{\it v}\}$ we will call $\alpha_\text{\it m}(\pi)$$=$$\alpha_\text{\it m}$$=$$\text{\it max}$ $\{|\alpha(\text{\it e}_1, \text{\it e}_2)|, |\alpha(\text{\it e}_2, \text{\it e}_3)|,\ldots, |\alpha(\text{\it e}_{\text{\it v}-1}, \text{\it e}_\text{\it v})|\}$ the maximum angle of alteration of the path. Given the value $\alpha_\text{\it m}: 0$$<$$\alpha_\text{\it m}$$<$$90$ an angle constrained path (ac-path) is such a path that $\alpha_\text{\it m}(\pi)\le\alpha_\text{\it m}$.

For reasons explained later in the paper, we are interested in a special type of any-angle and angle-constrained paths: $\Delta$-paths. A $\Delta$-path is a path (ac-path) in which every section is the $\Delta$-section with the possible exception of the final section. The paths depicted on the figure 1 are $\Delta$-paths, $\Delta$=5.

A potential solution for the MAPF problem is a set of partial potential solutions $\text{\it PS}$$=$$\{\text{\it PPS}^{(1)},\ldots, \text{\it PPS}^{(\text{\it n})}\}$, where a partial potential solution $\text{\it PPS}^{(\text{\it i})}$ is a tuple $\langle\pi^{(\text{\it i})}, \text{\it t}^{(\text{\it i})}\rangle, \pi^{(\text{\it i})}$ -- is a path (ac-path) for the {\it i}-th agent and $\text{\it t}^{(\text{\it i})}$ -- is the path's offset. We refer to a potential partial solution as a p-solution.

The path's offset is the period of time after the mission start that the agent waits on the ground prior to executing its path. Without loss of generality, assume that the speed of the agent is one grid cell-width per unit time. Thus $\text{\it t}^{(\text{\it i})}$ can now be measured in the same units as sections' and paths' lengths. Given that fact the cost of the MAPF solution is now the sum of all the path lengths and offsets: {\it cost}$(${\it PS}$)=\text{\it  cost}(\text{\it PPS}^{(1)})+\ldots+\text{\it cost}(\text{\it PPS}^{(\text{\it n})}), \text{\it cost}(\text{\it PPS}^{(\text{\it i})})=\text{\it t}^{(\text{\it i})}+\text{\it len}(\pi^{(\text{\it i})})$.
\begin{figure}[h]
\includegraphics[scale=0.66]{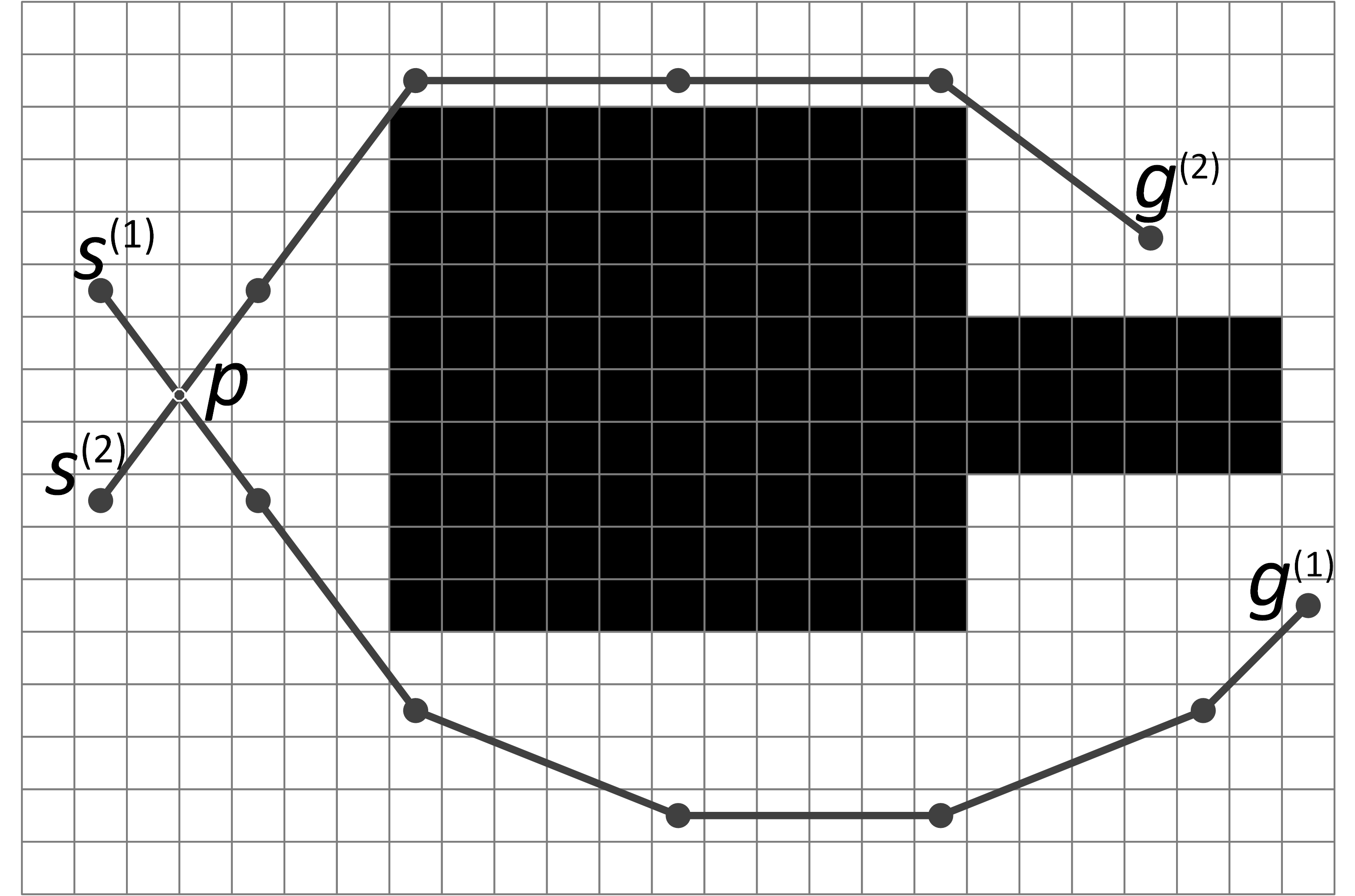}
\caption {Two paths on a grid. The upper path is the any-angle path, the lower path is the angle-constrained path ($\alpha_m=25^\circ$).}
\label{pic1}
\end{figure}

Suppose we are given the p-solution $\text{\it PPS}^{(\text{\it i})}$$=$$\langle\pi^{(\text{\it i})}, \text{\it t}^{(\text{\it i})}\rangle$ and the section $\text{\it e}_\text{\it j}^{(\text{\it i})}$$\in\pi^{(\text{\it i})}$. For the sake of simplicity, as we are talking now about single p-solution, upper indexes will be omitted. The g-value of section $\text{\it e}_\text{\it j}$ (in respect to {\it PPS}) is the following value: $\text{\it g}(\text{\it e}_\text{\it j}, \text{\it PPS}) = \text{\it t} + \text{\it len}(\text{\it e}_1) + \text{\it len}(\text{\it e}_2) + \ldots + \text{\it len}(\text{\it e}_{\text{\it j}-1})$. In other words {\it g}-value of the section is sum of two components: first is the offset as introduced in previous paragraph; second is the length of the partial path up to the considered section, or, from heuristic search community point of view, it's a $g$-value of section's startpoint. The g-value of a point lying on that section $\text{\it p}\in \text{\it e}_\text{\it j}$. {\it g} is given by $\text{\it g}(\text{\it p}, \text{\it e}_\text{\it j}, \text{\it PPS})=\text{\it g}(\text{\it e}_\text{\it j}, \text{\it PPS})+\text{\it dist}(\text{\it sp}(\text{\it e}_\text{\it j}), \text{\it p})$, where $\text{\it sp}(\text{\it e}_\text{\it j})$ is the startpoint of the section.

Consider two p-solutions, say for the first two agents, $\text{\it PPS}^{(1)}$ and $\text{\it PPS}^{(2)}$, and two sections $\text{\it e}_\text{\it i}^{(1)}\in\pi^{(1)}, \text{\it e}_\text{\it j}^{(2)}\in\pi^{(2)}$), again omitting the upper indices. If sections intersect or share some common part they considered to be in a potential conflict state: $\text{\it e}_\text{\it i}\cap \text{\it e}_\text{\it j}\ne\varnothing\Rightarrow (\text{\it e}_\text{\it i}, \text{\it e}_\text{\it j})\in \text{\it SPCON}$, where $\text{\it SPCON} \subseteq \text{\it E1} \times \text{\it E2}$ and {\it E1} is the set of the sections comprising $\pi^{(1)}$ and {\it E2} is the set of the sections comprising $\pi^{(2)}$.

Sections $\text{\it e}_\text{\it i}\in\pi^{(1)}$ and $\text{\it e}_\text{\it j}\in\pi^{(2)}$ are considered to be in a conflict state if they are in a potential conflict state and there exists a point belonging to both sections whose g-value are the same $\text{\it PPS}^{(1)}$ and $\text{\it PPS}^{(2)}$ are the same: $(\text{\it e}_\text{\it i}, \text{\it e}_\text{\it j}) \in \text{\it SCON}$ if $(\text{\it e}_\text{\it i}, \text{\it e}_\text{\it j}) \in \text{\it SPCON}$ and $\exists\text{\it p}$$:\text{\it p}\in \text{\it e}_\text{\it i} , \text{\it p}\in \text{\it e}_\text{\it j}, \text{\it g}(\text{\it p}, \text{\it e}_\text{\it i}, \text{\it PPS}^{(1)}) = \text{\it g}(\text{\it p}, \text{\it e}_\text{\it j}, \text{\it PPS}^{(2)})$.

Two p-solutions are considered to be in a conflict state if there exist at least one pair of sections (belonging to different p-solutions) which is in a conflict state: $(\text{\it PPS}^{(1)}, \text{\it PPS}^{(2)}) \in \text{\it CON}$ if $\exists \text{\it e}_\text{\it i}\in\pi^{(1)}, \text{\it e}_\text{\it j}\in\pi^{(2)}: (\text{\it e}_\text{\it i}, \text{\it e}_\text{\it j}) \in \text{\it SCON}$.

Given a set of {\it n} p-solutions, {\it e.g.} a potential solution {\it PS}, we call it to be conflict free if any two p-solutions from the set do not conflict: $\text{\it PS} \in \text{\it NoCON} \Leftrightarrow \nexists \text{\it PPS}^{(\text{\it i})}\in \text{\it PS}, \text{\it PPS}^{(\text{\it j})}\in \text{\it PS}: (\text{\it PPS}^{(\text{\it i})}, \text{\it PPS}^{(\text{\it j})}) \in \text{\it CON}$.

The goal of the multi-agent pathfinding task can now be formally stated as finding a conflict-free potential solution given the grid and the set of start and goal locations (grid cells) of the {\it n} agents. Cost of the solution is not subject to strict constraints (optimal solutions are not targeted) but low cost solutions are preferable.
\section{Approach overview}
The proposed algorithm uses a framework of decentralized initial path planning and centralized conflict resolution.

During path planning stage each of the agents plans it's path (any-angle or angle-constrained) independently and as a result the set of possibly conflicted solutions is produced. We rely on existing path finding algorithms, {\it e.g.} Theta* \cite{nash2007} for any-angle planning and LIAN \cite{yakovlev2015} for angle-constrained planning. We use a slight modification of Theta* that produces $\Delta$-paths. LIAN searches for $\Delta$-paths by default. Knowing that each constructed path has this special structure helps in eliminating conflicts.

In the conflict resolving stage the set of p-solutions obtained by the planner is refined to resolve all the conflicts. The resulting solution is not guaranteed to be optimal but our experiments show that the cost overhead is relatively small: less than 10\% or, in some cases, 1\%, for the scenarios we are interested in (100 of agents on a 501 x 501 grids being models of urban environment).

To resolve the conflicts we use two techniques. The first one is offset adjustment. Applying this technique  resolves the conflict by forcing one of the agents to wait before executing its plan, without changing the spatial path followed by either agent. The second technique is local re-planning which is done by constructing a small-scale detour out of $\Delta$-sections. As the path itself consists out of $\Delta$-sections the detour fits nicely.

We separate the given set of p-solutions into the two subsets (analogs of {\it OPEN} and {\it CLOSE} lists) and maintain a priority queue to gradually resolve the conflicts and guarantee the convergence of the algorithm.

\section {Resolving conflicts}

The basic core blocks of the conflict resolution algorithm are conflict identification and local detour computation. Therefore we describe these important subroutinues before describing the algorithm proper.

\subsection {Finding conflicts}

\begin{figure}[h]
\includegraphics[scale=0.66]{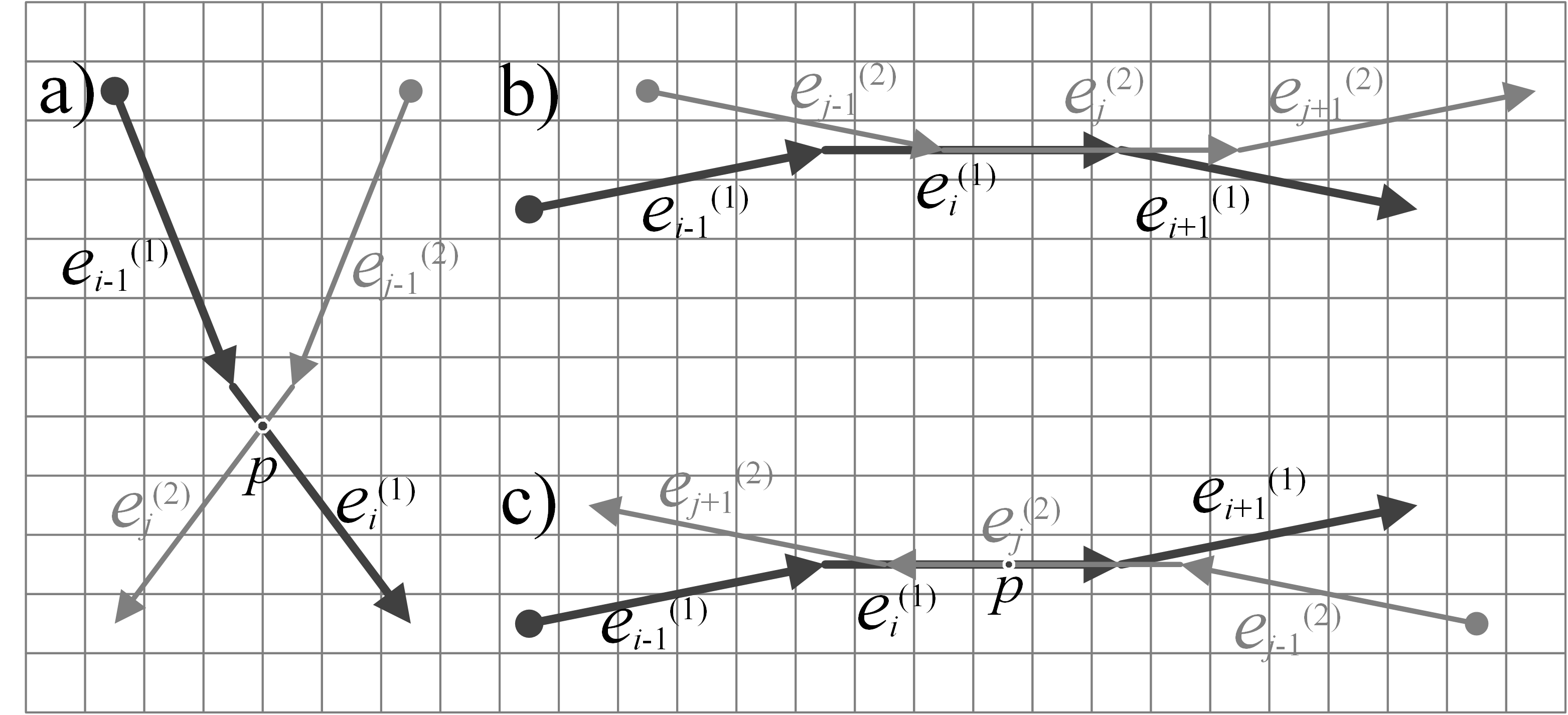}
\caption {Types of conflicts. a) Intersection conflict; b) Pursuit conflict; c) Head-on conflict.}
\label{pic2}
\end{figure}

Section conflicts can be classified into the following categories: intersection conflicts, pursuit conflicts, and head-on collisions (see figure ~\ref{pic2}). Conflicts of the first type appear when two sections are not collinear, intersect at one point and {\it g}-values of that intersection point calculated with the respect to both sections are equal. Pursuit conflicts appear when sections overlap and  point in the same direction. In that case there exists infinitely many points with equal g-values that belong to the overlap set (except the special case when the startpoint of one section is the endpoint of the other). Finally, head-on conflicts appear when the sections are anti-parallel and co-linear. It is easy to demonstrate that in this case there exists only one point from the overlap set for which the property of the equality of the {\it g}-values holds.

Consider now two p-solutions, say $\text{\it PPS}^{(1)}=\langle\pi^{(1)}, \text{\it t}^{(1)}\rangle,\\ \text{\it PPS}^{(2)}=\langle\pi^{(2)}, \text{\it t}^{(2)}\rangle$, and sections $\text{\it e}_\text{\it i}^{(1)}\in\pi^{(1)}, \text{\it e}_\text{\it j}^{(2)}\in\pi^{(2)}$ that are to be examined for a conflict. No conflict can occur if the distance between sections' endpoints exceeds the sum of their lengths. The same can be said about the {\it g}-values of their endpoints if the following condition does not hold: $\text{\it g}(\text{\it sp}(\text{\it e}_\text{\it i}^{(1)}), \text{\it PPS}^{(1)})\le \text{\it g}(\text{\it ep}(\text{\it e}_\text{\it j}^{(2)}), \text{\it PPS}^{(2)})$ and $\text{\it g}(\text{\it sp}(\text{\it e}_\text{\it j}^{(2)}), \text{\it PPS}^{(2)})\le \text{\it g}(\text{\it ep}(\text{\it e}_\text{\it i}^{(1)}), \text{\it PPS}^{(1)}),$ the conflict can not exist.

To identify a conflict after checking both abovementioned conditions and assuring the former can exist one should check (by applying well known computational geometry methods) whether the sections are collinear or not. If they are then either pursuit or head-on conflict should be considered, if not -- the intersection conflict should only be suspected.

If an intersection conflict is suspected one should compute the coordinates of the intersection point and then calculate the {\it g}-values of that point with respect to both p-solutions $\text{\it PPS}^{(1)}$ and $\text{\it PPS}^{(2)}$. Strictly speaking the difference between the calculated {\it g}-values should be zero to claim that sections are in a conflict state, but from practical point of view it is reasonable to use the following formula $|\text{\it g}(\text{\it p}, \text{\it e}_\text{\it i}, \text{\it PPS}^{(1)}) -  \text{\it g}(\text{\it p}, \text{\it e}_\text{\it j}, \text{\it PPS}^{(2)})| < r$: so-called {\it g}-equivalence condition.  Here {\it r} can be considered as the minimum safety radius and we suggest setting {\it r} to minimum grid distance, {\it e.g.} to the distance between centers of the horizontally (or vertically) adjacent cells.

To detect the pursuit conflict one need to verify that sections $\text{\it e}_\text{\it i}^{(1)}, \text{\it e}_\text{\it j}^{(2)}$ are collinear and co-directional. Then some point that belongs to the overlap set should be checked against {\it g}-values equivalence condition (as defined in the previous paragraph). The easiest way to choose such point is to take the startpoint of one of the sections involved in the potential conflict that belongs to the overlap set (there definitely be the one) -- see figure ~\ref{pic2}b.

Detection of the head-on is a more tricky task. It starts again with verifying that sections $\text{\it e}_\text{\it i}^{(1)}, \text{\it e}_\text{\it j}^{(2)}$ are collinear but opposite. Then the following conditions are checked: (1) $\text{\it g}(\text{\it sp}(\text{\it e}_\text{\it i}^{(1)}), \text{\it PPS}^{(1)}) + \text{\it dist}(\text{\it sp}(\text{\it e}_\text{\it i}^{(1)}), \text{\it ep}(\text{\it e}_\text{\it j}^{(2)})) \le \text{\it g}(\text{\it ep}(\text{\it e}_\text{\it j}^{(2)}), \text{\it PPS}^{(2)}); (2) \text{\it g}(\text{\it sp}(\text{\it e}_\text{\it j}^{(2)}), \text{\it PPS}^{(2)})+\text{\it dist}(\text{\it sp}(\text{\it e}_\text{\it j}^{(2)}), $ $\text{\it ep}(\text{\it e}_\text{\it i}^{(1)})) \le \text{\it g}(\text{\it ep}(\text{\it e}_\text{\it i}^{(1)}, \text{\it PPS}^{(1)})$. If both of them holds there exist a point belonging to the overlap set that satisfies the g-values equivalence condition, meaning the conflict does exist.

The conflict detection procedures can now be combined into the {\it FindFirstConflict} routine that later becomes one of the main building blocks of the proposed conflict resolution algorithm. This function takes a p-solution $\text{\it PPS}^{(\text{\it i})}\in \text{\it PS}$ and a set of p-solutions $\text{\it PS}'\subset \text{\it PS}$ as an input and returns the first conflict of PPS as an ouput, {\it e.g.} a pair of sections $\text{\it e}_\text{\it v}^{(\text{\it i})}, \text{\it e}_\text{\it w}^{(\text{\it k})}$$: \text{\it e}_\text{\it v}^{(\text{\it i})}\in\pi^{(\text{\it i})}, \text{\it e}_\text{\it w}^{(\text{\it k})}\in\pi^{(\text{\it k})}, \text{\it PPS}^{(\text{\it k})}\in \text{\it PS}', (\text{\it e}_\text{\it v}^{(\text{\it i})}, \text{\it e}_\text{\it w}^{(\text{\it k})})\in \text{\it SCON}$  and $\nexists \text{\it t}, \text{\it e}_\text{\it q}^{(\text{\it z})}$$\in \text{\it PS}'$$:  \text{\it e}_\text{\it v}^{(\text{\it i})}\in\pi^{(\text{\it i})}, \text{\it t}<\text{\it v}, (\text{\it e}_\text{\it t}^{(\text{\it i})}, \text{\it e}_\text{\it q}^{(\text{\it z})})\in \text{\it SCON}$. Technically this is done by iterating over the sections comprising $\pi^{(\text{\it i})}$ from the first to the last one and performing the intersection, pursuit and head-on inspections as described above. If no conflicts are found the function returns an empty set.

\subsection {Planning local detours}

Local re-planning is one of two techniques which is used to eliminate conflicts between p-solutions forming a potential solution of MAPF problem (the other one is simply incrementing the p-solution offset). Consider two p-solutions $\text{\it PPS}^{(1)}=\langle\pi^{(1)}, \text{\it t}^{(1)}\rangle, \text{\it PPS}^{(2)}=\langle\pi^{(2)}, \text{\it t}^{(2)}\rangle$ and two sections $\text{\it e}_\text{\it i}^{(1)}\in\pi^{(1)}, \text{\it e}_\text{\it j}^{(2)}\in\pi^{(2)}$, that are
known to conflict are given as well as the information on which path (either $\pi^{(1)}$ or $\pi^{(2)})$ should be altered. Let it be $\pi^{(2)}$ in our case. We will subsequently ommit upper indices, which are assumed to be 2.

\begin{figure}[h]
\center{\includegraphics[scale=0.8]{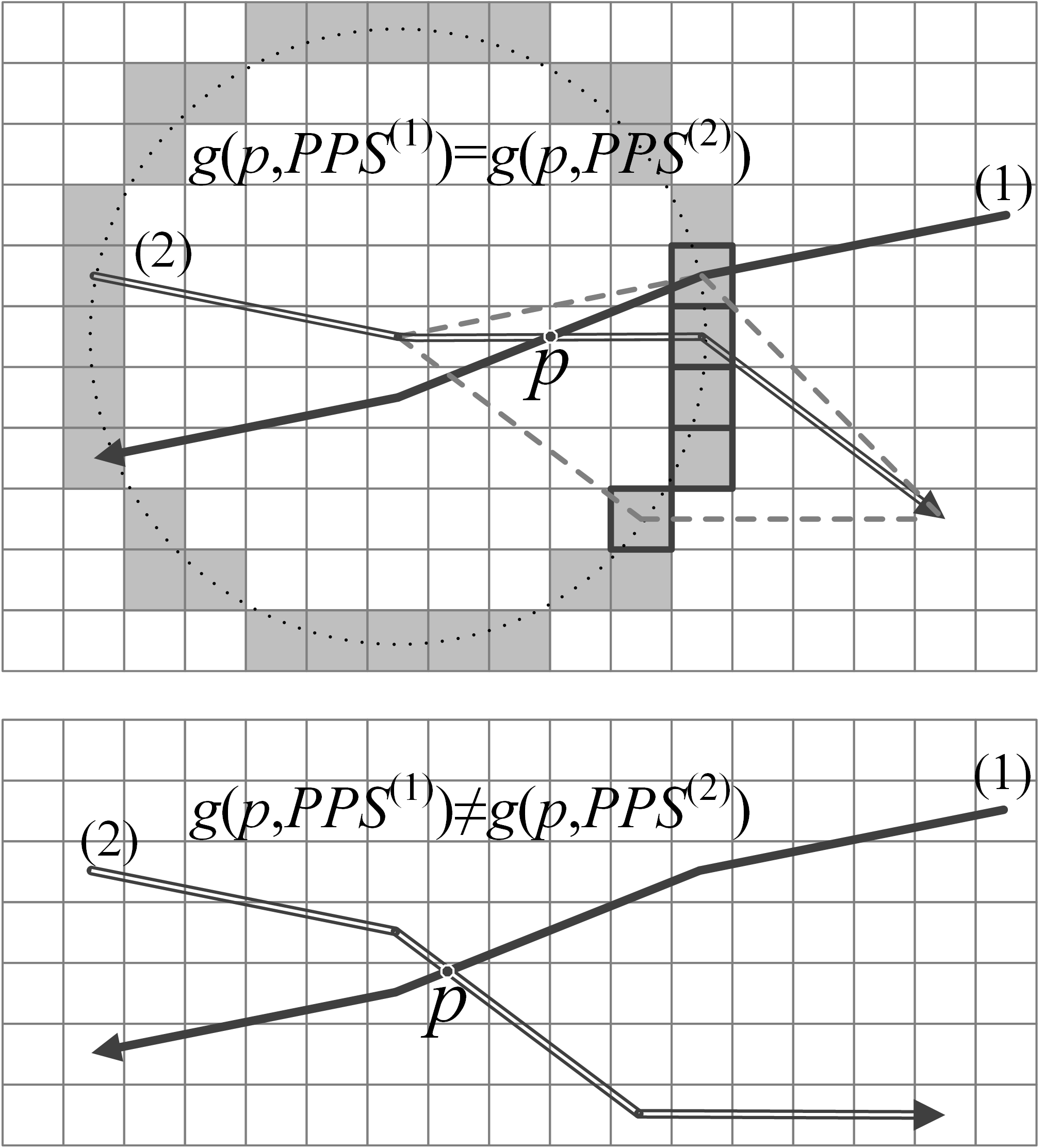}}
\caption {Local re-planning via constructing a detour. a) Cells identified by the Midpoint algorithm are shaded grey. Cells that are left in the candidates set after pruning are highlighted in bold. b) p-solutions after re-planning.}
\label{pic3}
\end{figure}

The idea of local re-planning is to construct a slight detour by changing the sections $\text{\it e}_\text{\it j}$ and $\text{\it e}_{\text{\it j}+1}$. Given that the paths are $\Delta$-paths the detour can be constructed in the following manner. - see figure  ~\ref{pic3}.

First the cells lying on the circle of radius $\Delta$ with the center in $\text{\it sp}(\text{\it e}_\text{\it j})$ are identified (by the midpoint algorithm \cite{pitteway1985}). These cells will be referred to as {\it CIRCLE}. Then all $\text{\it a}_{\text{\it kl}}\in CIRCLE$ that do not satisfy the angle of alteration constraint: $|\alpha(\text{\it e}_{\text{\it j}-1}, \text{\it e}_{\text{\it new}1})|>\alpha_\text{\it m}, \text{\it e}_{\text{\it new}1}=\langle \text{\it ep}(\text{\it e}_{\text{\it j}-1}), \text{\it a}_{\text{\it kl}}\rangle$ are pruned. Here $\alpha_\text{\it m}: 0<\alpha_m<90$ is the input parameter (in case angle-constrained path is under consideration it should be equal to the one that was used to find such path). Pruning helps to avoid quirky paths with sharp heading changes in the opposite directionsand also ensures that ac-paths continue to satisfy their angle of alteration constraint (at least on the \{$\text{\it e}_{\text{\it j}-1}, \text{\it e}_{\text{\it new}1}$\} path segment).

After the set of the candidates is pruned, the cell $\text{\it a}_{\text{\it kl}}\in \text{\it CIRCLE}$ such that both sections $\text{\it e}_{\text{\it new}1}=\langle \text{\it ep}(\text{\it e}_{\text{\it j}-1}), \text{\it a}_{\text{\it kl}}\rangle$ and $\text{\it e}_{\text{\it new}2}=\langle \text{\it a}_{\text{\it kl}}, \text{\it ep}(\text{\it e}_{\text{\it j}+1})\rangle$ are traversable, is chosen to form a detour segment (in case ac-path is altered the cell $\text{\it a}_\text{\it kl}$ must also satisfy $|\alpha(\text{\it e}_{\text{\it new}1}, \text{\it e}_{\text{\it new}2},)|\le \alpha_\text{\it m}$ and $|\alpha(\text{\it e}_{\text{\it new}2}, \text{\it e}_{\text{\it j}+2},)|\le\alpha_\text{\it m}$ conditions). In general $\text{\it a}_{\text{\it kl}}$ selection can be done arbitrary. In this paper we recommend chosing the cell that maximizes the detour deviation angle, {\it e.g.} $\text{\it a}_{\text{\it kl}} = \text{\it argmax}_{\text{\it a}_{\text{\it kl}}\in \text{\it CIRCLE}} |\alpha(\text{\it e}_{\text{\it j}-1}, \langle \text{\it a}_{\text{\it kl}}, \text{\it ep}(\text{\it e}_{\text{\it j}+1})\rangle)|$.

\subsection {Conflicts resolution algorithm}

With the major subroutinues introduced, we can now describe the conflict resolution algorithm itself.
	
The input of the algorithm is a 4-tuple $\langle${\bf \emph {PS}}, $\bm{ \Delta, \alpha_m}$, {\bf \emph {wait}}$\rangle$, where {\bf PS} is the initial potential solution (set of any-angle or angle-constrained paths provided by the path finder combined with zero offsets),  $\bm{\Delta}$ is the section size that was used during path finding and that will be used for local re-planning, $\bm{\alpha_m}$ is a parameter for {\it ComputeLocalDetour} and {\it {\bf wait}} is the offset increment (in case the conflict can not be eliminated via re-planning, corresponding p-solution execution will be delayed for further {\bf wait} timestamps).
\begin{algorithm}
\begin{scriptsize}
\SetNlSty{text}{}{:}
\IncMargin{1em}
\caption{Conflicts resolution}\label{euclid}
{\small \KwIn{\bf \emph{PS}, $\bm{\Delta, \alpha_m}$, \bf \emph{wait}; {\bf Output: \emph{PS}}$'$$\in$NoCON}}
\{{\it HEAD, TAIL}\}$\leftarrow${\it FormHeadAndTail}({\bf \emph{PS}})\\
\While {{\it TAIL}$ \ne$$\varnothing$}
{
	$\text{\it PPS}^{\text{({\it cur})}}$$=$argmin$_{\text{\it PPS}\in \text{\it TAIL}}${\it NumberOfConflicts(PPS, TAIL$\cup$HEAD)}\\
	{\it TAIL}.remove($\text{\it PPS}^{\text{({\it cur})}}$)\\
	\While{(\{$\text{\it e}_\text{\it v}^\text{({\it cur})},\text{\it e}_\text{\it w}^\text{({\it k})}$\}$\leftarrow${\it FindFirstConflict(}$\text{\it PPS}^\text{({\it cur})}$,HEAD))$\ne$$\varnothing$}
	{
		$\pi^{\text{({\it new})}}$$\leftarrow${\it ComputeLocalDetour}$(\text{\it PPS}^\text{(\it{cur})}$,$\text{\it e}_\text{\it v}^\text{(\it{cur})}$,$\text{\it PPS}^\text{(\it{k})}$,$\bm{\Delta}$,$\bm{\alpha_m})$\\
		\eIf{$\pi^\text{({\it new})}=\pi^\text{({\it cur})}$}
		{
			$\text{\it t}^\text{({\it cur})}+$$={}${\bf \emph {wait}}
		}
		{
			$\{\text{\it e}_\text{\it t}^\text{({\it new})},\text{\it e}_\text{\it s}^\text{({\it m})}\}$$\leftarrow${\it FindFirstConflict}$\text{\it (PPS}^\text{({\it new})}$,{\it HEAD})\\
			\eIf {\{$\text{\it e}_\text{\it t}^\text{({\it new})},\text{\it e}_\text{\it s}^\text{({\it m})}\}=\varnothing$ {\bf or} {\it t$>$v}}
			{
				$\pi^\text{({\it cur})}=\pi^\text{({\it new})}$
			}
			{
				$\text{\it t}^\text{({\it cur})}+$$={}${\bf \emph {wait}}
			}
		}
		
	}
	{\it HEAD}.add$\text{\it (PPS}^\text{({\it cur})})$
}
{\bf return} {\it HEAD}
\end{scriptsize}
\end{algorithm}

The main idea of the proposed algorithm is to split the initial solution into two disjoint subsets of p-solutions {\it HEAD} and {\it TAIL} (analogs of {\it CLOSE} and {\it OPEN} lists well known to heuristic search community) and maintain them in such a way that {\it HEAD} always contains p-solutions that do not conflict with each other (so-called {\it HEAD} condition) and {\it TAIL} contains the conflicting p-solutions. Assuring this property always holds, algorithm removes p-solutions from {\it TAIL} one by one, refines them to satisfy {\it HEAD} condition and adds them to {\it HEAD}. So on each iteration {\it TAIL} shrinks per one element and {\it HEAD} enlarges. After a finite number of iterations {\it TAIL} will be empty and {\it HEAD} will contain all the p-solutions. The problem is then solved.

The first step of the algorithm is forming {\it HEAD} and {\it TAIL}. Technically this is done in the following way (pseudocode is omitted for the sake of space). The provided p-solutions are examined and those free of conflicts are placed in{\it  HEAD}, while those containing conflicts are placed in {\it TAIL}. Then the algorithm tests each p-solution in the {\it TAIL} list.  If it does not conflict with any p-solution in the {\it HEAD} list, it is added to the {\it HEAD} list. 

After {\it HEAD} and {\it TAIL} are initialized the algorithm enters the main loop (lines 2-15). Each iteration starts from choosing the p-solution with highest priority (lowest number of p-solutions it's in conflict with) from {\it TAIL} -- $\text{\it PPS}^{(\text{\it cur})}$. Then it is removed from {\it TAIL}, refined and added to {\it HEAD}. The core step of the main loop is the p-solution refinement (lines 5-14).

Refinement is the iterative proccess of adjusting the current p-solution, $\text{\it PPS}^{(\text{\it cur})}$$=$$\langle\pi^{(\text{\it cur})}, \text{\it t}^{(\text{\it cur})}\rangle$, until it has no conflict with any p-solution from the {\it HEAD} set. To check this the function {\it FindFirstConflict} (as described  above) is used. If no conflicts exist it returns an empty set. If the conflicts does exist it returns the first conflict of $\pi^{(\text{\it cur})}$. Retrieving the first conflict is important and helps to avoid cycles (will be discussed shortly on).

After the conflict is identified the $\text{\it PPS}^{(\text{\it cur})}$ is altered in the following way. First, an attempt is made to construct a local detour by invoking the {\it ComputeLocalDetour} function. If the attempt fails  the algorithm increments $\text{\it t}^{(\text{\it cur})}$.If the local detour is successfully found one needs to ascertain that the conflict did not shift backwards in time  (line 11) which would be unacceptable as it may introduce loops in the resolution algorithm. If this constraint is violated, the path modification is reverted and the conflict is resolved by incrementing the $\text{\it PPS}^{(\text{\it cur})}$ offset instead. Note that incrementing the offset is a cycle safe operation as it always results in a one-way non-backtrackable alteration of the p-solution set.

\subsection {Comparison with CBS algorithm}

One can think of the introduced algorithm as of the greedy modification of the CBS algorithm \cite{sharon2015}, \cite{boyarski2015a}, \cite{boyarski2015b}. CBS is the two-leveled method, relying on optimal path planning for an individual agent at the low level and on a specific conflict resolution strategy at the high level. CBS can potentially explore all the variants of conflicts elimination and backtrack, and thus guarantees optimality. Our algorithm never backtracks -- once a p-solution is considered to satisfy {\it HEAD} condition it's fixed for good. That leads to sub-optimal solutions but speeds up the search.

In fact, solving MAPF problem optimally under any-angle assumption is a non-trivial task. Of course, CBS can use an any-angle planner as its underlying planner but only in case it guarantees handling the CBS-constraints. The latter come in the form of either $\langle \text{\it a}_\text{\it i}, \text{\it v}_\text{\it j}, \text{\it t}_\text{\it k} \rangle$ or $\langle \text{\it a}_\text{\it i}, \text{\it e}_\text{\it j}, \text{\it t}_\text{\it k} \rangle$, {\it e.g.} agent $\text{\it a}_\text{\it i}$ is disallowed to occupy vertex $\text{\it v}_\text{\it j}$ or edge $\text{\it e}_\text{\it j}$ at the timestamp $\text{\it t}_\text{\it k}$. These constraints are identified by CBS at the high level and then passed to low level planner. As any-angle conflicts are not strictly tied to grid vertices a specific method of vertex-based constraints identification should be proposed or such conflicts should be discarded and only edge-based conflicts should be considered. Handling edge conflicts is a tricky task as any-angle planners construct edges on-the-fly during the search and, in general, an edge can be formed out of any pair of grid cells. This can lead to constructing an edge that formally satisfies CBS-edge constraint but does not eliminate the conflict. Consider a newly generated edge that is not a precise duplicate of $\text{\it e}_\text{\it j}$ but slightly differs in its endpoints. Chances are following this edge still leads agent $\text{\it a}_\text{\it i}$ to collision with the same other agent. So one should guarantee that not only the prohibited edge is omitted in the newly constructed path but that the source conflict is really resolved. Tackling these issues is an appealing direction of research but goes beyond the scope of the paper.
\section {Experimental evaluation}

The hardware setup for the experimental evaluation was the Windows7-operated PC, iCore2 quad 2.5GHz, 2Gb RAM. All the algorithms: Theta*, LIAN and proposed conflict resolution algorithm were coded in C++ using the same data structures and programming techniques. Source code is available at www.pathplanning.ru/public/wompf-2016. Urban outdoor navigation scenario was targeted and cooperative path finding for a group of 100 unmanned aerial vehicles performing nap-of-the-earth flight was addressed.

100 grids were involved in the experiments. Each of them was automatically constructed using the data from the OpenStreetMaps (OSM) database \cite{haklay2008}. To generate a grid a 1347m x 1347m fragment of actual city environment was retrieved from OSM and discretized to 501 x 501 grid. Cells corresponding to the areas occupied by buildings were marked un-traversable. 20-25\% of un-traversable cells were present on each grid.

Two different techniques of start and goal allocation were implemented, resulting in two sub-collections (type-1 and type-2).). In the first type of allocation 100 start cells were chosen randomly close to grid borders, {\it e.g.} outside the region bounded by the 2 horizontal and 2 vertical imaginary lines that lie parallel to grid borders 50 cells away from them. Goal cells were chosen respectively from the opposite border region. For the allocation of second type a square region sized 50$\times$50 was identified on each grid lying close to a grid border and all the 100 start cells were chosen randomly from that region. All goal cells were chosen from the similar 50$\times$50 region lying on the opposite side of the grid. Two tasks per grid per sub-collection were generated, thus resulting in 400 test cases.

For each task in a testbed 1) $\bm{\Delta}$-paths were found using Theta* and LIAN algorithms (the input parameters were set to be as follows: $\bm{\Delta}$$=$5, $\bm{\alpha_m}$$=$25 and {\bf \emph{wait}}$=$5), and 2) the existing conflicts were eliminated by the proposed algorithm.

Table 1 contains the information on conflicts distribution after the first step (all figures are the averaged ones).
\begin{table}[H]
\caption{\label{table1} Conflicts distribution after path finding.}

\begin{tabular}{  c | c c | c  c  }
\hline
	 & \multicolumn{2}{|c|} {\bf Theta*} & \multicolumn{2}{|c}  {\bf LIAN}  \\ 
	 & \multicolumn{1}{|c}{\bf Type-1} & {\bf Type-2} & {\bf Type-1} & {\bf Type-2} \\ \hline
	{AgentConflicts} & {42.3} & {94.83} & 38.63 & 98.25 \\
	{SectionConflicts} &  {76.15} & {2423.56} & {64.92} & {2457.41} \\ \hline

\end{tabular}
\end{table}

In presented table ``SectionConflicts'' stands for number of conflicts as defined in the paper and ``AgentConflicts'' means the number of agents which paths have at least one conflict with other agents' paths. As one can see, even in case agents are sparsely distributed over the grid (type-1), conflicts occur quite often and nearly 40\% of agents do have conflicts. In case agents initial and target locations are concentrated on quite a bounded area (type-2), number of conflicts significantly increases, up to 95\% and more. Large gap between AgentConflicts and SectionConflicts for type-2 tasks speaks for the fact that each agent a) is in a conflict state with more than one other agent (actually 2-5 as additional analysis of collected data shows), b) many coupled conflicts exist, {\it e.g.} a small number of agents p-solutions (paths) produce large number of conflicts between each other.

Resolution of the obtained conflicts by the proposed algorithm leads to the following results -- see Table 2.
\begin{table}[H]
\caption{\label{table2} Conflicts resolution stats.}

\begin{tabular}{  c | c  c | c  c  }
\hline
	  & \multicolumn{2}{|c|} {\bf Theta*} & \multicolumn{2}{|c}  {\bf LIAN}  \\ 
	 & \multicolumn{1}{|c}{\bf Type-1} & {\bf Type-2} & {\bf Type-1} & {\bf Type-2} \\ \hline
	  \multicolumn{5}{c} {\it Agents}  \\ \hline
	Delayed & 12.07 & 64.685 & 15.3 & 72.335 \\ 
	Replanned & 14.73 & 58.565 & 7.77 & 64.09 \\ 
	Unchanged & 79.54 & 34.33 & 80.56 & 26.44 \\ \hline
	 \multicolumn{5}{c}{\it Attempts to resolve SectionConflicts by}  \\ \hline
	PathOffset & 21.025 & 746.085 & 26.575 & 1176.19 \\ 
	Replan & 30.935 & 1442.185 & 12.7 & 638.3 \\ \hline
\end{tabular}
\end{table}

As one can see in order to eliminate all the conflicts for type-1 scenarious only 1/2 of the conflicting p-solutions should be altered (solutions for 20 agents were changed whilst 40 were in a conflict state initially). For type-2 tasks this value is much lower 1/4 -- 1/3 depending on the path finding algorithm. It also can be stated that re-planning strategy is much more successful if non angle-constrained paths are considered. Setting the p-solution's offset (delaying the agent) seems to be more efficient for solving type-2 tasks rather than type-1.

Noteworthy are the figures that express how many conflicts were tried to be resolved (last two rows of the table). As one can see for type-2 tasks way to many re-planning attempts were made. Obviously the majority of these attempts fail to produce a conflict-free p-solution (see the row \#2). We suggest that a conflict after local re-planning in many cases just ``moved forward'' resulting in a sequence of unsuccessful re-planning attempts. So it is a good idea to handle such resolution patterns in future.
Runtimes and solution costs are presented in Table 3.
\begin{table}[H]
\caption{\label{table3} Average runtimes and solution costs. PF -- path finding; CR -- conflict resolution.}
\begin{tabular}{ c  c | c  c | c  c  }
\hline
	 & & \multicolumn{2}{|c|} {\bf Theta*} & \multicolumn{2}{|c}  {\bf LIAN}  \\ 
	& & {\bf Type-1} & {\bf Type-2} & {\bf Type-1} & {\bf Type-2} \\ \hline \rule{0cm}{3mm}
	\multirow{2}{*}{PF} & Time(s) & 7.3783 & 6.9426 & 9.5827 & 5.387 \\
	 & Cost & 46926 & 46722 & 49636 & 49651 \\
	\multirow{3}{*}{CR} & Time(s) & 0.1466 & 0.804 & 0.1441 & 0.5926 \\
	& \multirow{2}{*}{Cost} & 47034  &50477& 49772  & 55566  \\ 
	& & +0.23\% & +8.04\% & +0.27\% & +11.91\% \\ \hline
\end{tabular}
\end{table}
Having these results one can claim that the proposed conflict resolution algorithm is characterized by low runtime and minor increase in solution cost (see the last row) and thus can be used in practice to solve MAPF problems when any-angle or angle-constrained path are to be constructed.

Obtained results leave a question: does the suggested approach work with more agents? To address this question and to estimate the limits of the introduced conflicts resolution method we have conducted another series of the experiments involving 200-500-1000-2000-4000 agents (solving MAPF for more than 4000 agents took us too much time and thus can be considered as infeasible within suggested framework).

Theta* algorithm was used as a path planner and type-1 start-goal allocation strategy was considered. As both path planning and conflicts resolution slow down significantly as the number of agents increase, only 10 maps out of 100 available were used to run the experiments. The input parameters for the conflict resolution algorithm were set to be as follows: $\bm{\Delta}$$=$5, $\bm{\alpha_m}$$=$25 and {\bf \emph {wait}}$=$5.
Table 4 contains the information on conflicts distribution after path planning (all figures are the averaged ones).

\begin{table}[H]
\caption{\label{table4}Number of conflicts after path finding for 200-4000 agents.}

\begin{tabular}{  c | c  c  c  c  c  }
\hline \rule{0cm}{4mm}
	 & {\bf 200} & {\bf 500} & {\bf 1000} & {\bf 2000} & {\bf 4000}  \\ \hline \rule{0cm}{3mm}
	{AgentConflicts} & {120} & {427} & 962 & 1988 & 3997 \\
	{SectionConflicts} & 327 &  {2131} & {8674} & {35217} & {145551} \\ \hline

\end{tabular}
\end{table}

Presented figures support semi-obvious claim that the number of conflicts grows significantly as the number of agents increases. If the latter exceeds 1000 it's likely that almost every agent is involved in some conflict(s) (see the first row of the table). As for the conflicts themselves, one can propose that the number of conflicts exhibits  quadratic growth in the number of agents (see ``SectionConflicts'' row). This appropriately means that the conflicts are getting more coupled, {\it e.g.} for 200 agents the ratio ``SectionConflicts / AgentConflicts'' equals 1.63, meaning that each agent has conflicts with 1.6 other agents on average, the same ratio for 2000 agents is 17.6, for 4000 -- 36.4.

Resolution of the obtained conflicts leads to the following observations -- see Table 5.
\begin{table}[H]
\caption{\label{table5} Conflicts distribution after path finding.}
\begin{tabular}{  c | c  c  c  c  c  }
\hline
\rule{0cm}{4mm}
	  & {\bf 200} & {\bf 500} & {\bf 1000} & {\bf 2000} & {\bf 4000} \\ \hline 
	  \multicolumn{6}{c} {\it Agents}  \\ \hline
\rule{0cm}{3mm}
	Delayed & 40 & 191 & 543 & 1395 & 3258 \\ 
	Replanned & 43 & 200 & 557 & 1409 & 3272 \\ 
	Unchanged & 142 & 261 & 390 & 535 & 680 \\ \hline 
	 \multicolumn{6}{c}{\it Attempts to resolve SectionConflicts by}  \\ \hline
\rule{0cm}{3mm}
	PathOffset & 86 & 815 & 5162 & 32757 & 176158 \\ 
	Replan & 123 & 1158 & 5835 & 28557 & 128253 \\ \hline
\end{tabular}
\end{table}

As number of agents grows, number of unchanged p-solutions shrinks (see the third row). For 1000 agents  61\% of p-solutions were altered, for 4000 -- 83\%, while it's only 29\% for 200. Analyzing the number of delayed and re-planned agents one can say that as the number of agents grows, chance of resolving a distinct p-solution conflict by only one technique, either local re-planning or setting the time offset, diminishes. It's 43\% percent of p-solutions which were both re-planned and delayed for 200 agents present, and 96\% for 2000. These are calculated using the formula: ({\it n $-$ d $-$ r})$/${\it n}, where {\it n} is the number of altered p-solutions (Total -- ``Unchanged''), {\it d} is the number of delayed-only p-solutions (``Delayed'') and {\it r} is the number of replanned p-solutions (``Replanned'').

An interesting observation can be made if one takes a deeper look at the number of conflicts that were actually tried to be resolved (last two rows of the table). Consider a value ({\it po$+$r})$/${\it sc}, where {\it po} is ``PathOffset'', {\it r} is ``Replan'' and {\it sc} is ``SectionConflicts'' (from table 4), which characterizes the ``attempt/success'' ratio. This value for 200-500-1000-2000-4000 agents is: 0.64-0.93-1.27-1.74-2.1. That means, in case more than 1000 p-solutions are under consideration, solving a conflict leads to creating another one, so the total number of conflicts actually doubles up. This obviously has a great impact on the algorithm runtime -- see table 6.

\begin{table}[H]
\caption{\label{table6}Average runtimes and solution costs' overheads after conflict resolution.}

\begin{tabular}{  c | c  c  c  c  c  }
\hline
\rule{0cm}{4mm}
	 & {\bf 200} & {\bf 500} & {\bf 1000} & {\bf 2000} & {\bf 4000}  \\ \hline 
\rule{0cm}{3mm}
	{Time(s)} & {0.62} & {6.19} & 49.69 & 466.8 & 5072.42 \\
	{\specialcell{Cost \\ overhead}} & 0.47\%& 1.78\% &  {5.61\%} & {17.74\%} & {47.59\%} \\ \hline

\end{tabular}
\end{table}

Runtime of the algorithm grows dramatically as the number of agents (and conflicts) enlarges. According to the figures observed, one can talk about quadratic dependency. At the same time, solution cost overhead tends to showcase  linear growth. As for the absolute figures, it's very hard to comment on them without a comparison with other algorithms, which are not present nowadays for the any-angle MAPF domain (to the best of our knowledge). By now we can only suggest that from practical point of view the reasonable limit of the proposed approach (within the considered navigation scenario) is 1000 of agents per grid.

\section{Summary}
In this work we have studied a multi-agent path finding problem in case square grids are used as the environment model and each agent is allowed to move into arbitrary directions. We have proposed procedures of conflict identification and conflict elimination via local detour planning. These procedures have been used to present the original conflicts resolution algorithm that can be a part of any MAPF framework relying upon independent path finding and centralized resolving of conflicts. Although the algorithm is not guaranteed to be optimal, experimental evaluation showed that it's application to practical tasks leads to low solution cost overheads (at least for the scenarios we were interested in).
	
In future we intend to perform the theoretical studies of the algorithm in order to reveal it's properties and to prove that it is complete. Another direction of research is the modification of local re-planning techniques aimed at resolving ``conflict moving forward'' problem.

\section*{Acknowledgements}
We thank Glenn Wagner for the valuable suggestions and his help in improving the paper.

\bibliographystyle{named}
\bibliography{ijcai16}

\end{document}